\documentclass{article}
\usepackage{spconf,amsmath,graphicx}

\usepackage[utf8]{inputenc}
\usepackage{xspace}
\usepackage{caption}
\usepackage{subcaption}
\usepackage{multirow}
\usepackage{todonotes}
\usepackage{soul}
\usepackage[hidelinks]{hyperref}
\usepackage[sort&compress,noabbrev,capitalise,nameinlink]{cleveref}

\makeatletter
\DeclareRobustCommand\onedot{\futurelet\@let@token\@onedot}
\def\@onedot{\ifx\@let@token.\else.\null\fi\xspace}

\def\eg{e.g\onedot} 
\def\ie{i.e\onedot} 
\def\cf{c.f\onedot} 
\def\etc{etc\onedot}

\makeatother

\usepackage{amssymb}
\usepackage{pifont}
\newcommand{\cmark}{\ding{51}}%
\newcommand{\xmark}{\ding{55}}%

\title{Self-SuperFlow: Self-supervised Scene Flow Prediction in Stereo Sequences}
\name{Katharina Bendig$^{1}$, René Schuster$^{2}$, Didier Stricker$^{1,2}$
\thanks{This work was partially funded by the Federal Ministry of Education and Research Germany under the project DECODE (01IW21001).}
}
\address{
$^{1}$Technische Universität Kaiserslautern\\
$^{2}$DFKI -- German Research Center for Artificial Intelligence\\
\texttt{firstname.lastname@dfki.de}
}
\begin{document}
%
\maketitle
\begin{abstract}
In recent years, deep neural networks showed their exceeding capabilities in addressing many computer vision tasks including scene flow prediction. However, most of the advances are dependent on the availability of a vast amount of dense per pixel ground truth annotations, which are very difficult to obtain for real life scenarios. Therefore, synthetic data is often relied upon for supervision, resulting in a representation gap between the training and test data. Even though a great quantity of unlabeled real world data is available, there is a huge lack in self-supervised methods for scene flow prediction. Hence, we explore the extension of a self-supervised loss based on the Census transform and occlusion-aware bidirectional displacements for the problem of scene flow prediction. Regarding the KITTI scene flow benchmark, our method outperforms the corresponding supervised pre-training of the same network and shows improved generalization capabilities while achieving much faster convergence. 
\end{abstract}

\section{Introduction} \label{sec:intro}


The safe navigation of robots and autonomous vehicles strongly depends on the sufficient perception of their surroundings. This especially necessitates the correct estimation of the motion of other traffic participants. Scene flow is one way to represent the rich perceptual information needed as it depicts the complete 3D motion field of objects in the environment and does not introduce additional ambiguity compared to its projection on the image plane (optical flow).

A currently famous approach for scene flow prediction is the usage of convolutional neural networks since they are able to outperform many traditional methods \cite{fischer2015flownet}. These methods are trained in a supervised manner and thus rely heavily on the availability of a large amount of ground truth data, which is difficult to obtain in a real world setting, especially for the problem of scene flow prediction. For that reason, synthetically generated images are often used for pre-training, since they allow an easy generation of dense ground truth scene flow. However, synthetic images display intrinsic differences to the real world images introducing an inherent representation gap between the training and test data sets. Real world data is often only utilized for an optional fine-tuning stage after the synthetic pre-training. This results in a reduced generalization for real world scenarios and thus makes the application in areas like autonomous driving challenging.

One way to address these obstacles is the utilization of self-supervision, \ie the network obtains its training signal (loss) based on the available input data. However, there is currently a great lack in self-supervised approaches for stereo scene flow prediction. Therefore, we propose a novel method to solve scene flow prediction in this self-supervised setting for the first time. Our innovative strategy infers multiple displacements based on the availability of rich information due to the two stereo image pairs used. These displacements can be applied for occlusion reasoning as well as bidirectional reconstruction losses for arbitrary reference images. Our approach further utilizes a ternary census loss following the example of \cite{meister2018unflow} and explores the effect of optimization techniques for self-supervision like cost volume normalization introduced by \cite{jonschkowski2020matters}.
Our novel self-supervised training method for stereo scene flow prediction is called Self-SuperFlow.
On the KITTI evaluation data set \cite{Geiger2013IJRR, Menze2015CVPR}, it vastly outperforms the equivalent supervised pre-training of the same network with 20 percentage points less outliers, while drastically cutting down the convergence time of the training.

\section{Related Work} \label{sec:relwork}
\subsection{Scene Flow Prediction}
Optical flow approaches build the foundation for many scene flow prediction methods. Thus traditional approaches are often based on variational methods, which involve the minimization of an energy function \cite{conf/icra/HerbstRF13, Basha2010MultiviewSF}. However, due to their iterative nature, long run time, and constancy assumptions, variational methods are non applicable in real world scenarios. Another strategy is to assume rigidity, like \eg in \cite{Menze2015CVPR,6751281}, albeit objects in real world settings often do not move rigidly thus breaking the assumption. Therefore, the preferred method nowadays is the usage of neural networks, which are trained in a supervised manner \cite{fischer2015flownet, saxena2019pwoc3d}. However, this depends strongly on the availability of a large amount of ground truth data, which is difficult to obtain for a real world setting. In order to overcome this dependency, we instead propose the application of self-supervised training, which is able to leverage the vast amount of unlabelled real world data. 

\subsection{Self-Supervised Optical Flow} 
Self-supervised methods for optical flow prediction are mostly based on the photometric similarity of warped frames \cite{ahmadi2016unsupervised,jason2016back, zhu2017guided}. However, one obstacle in the usage of the photometric loss is its lack of invariance towards monotonic illumination changes, which appear commonly in real world scenarios. The authors of \cite{meister2018unflow} approach this issue by utilizing 
a ternary census loss. They further leverage a bidirectional flow setup for occlusion detection. Moreover, approaches like \cite{alletto2017transflow, yin2018geonet} apply geometric constraints or reason about the rigidity of objects in the scene. 
General improvements for self-supervision in optical flow settings are described in \cite{jonschkowski2020matters} and involve among other things cost volume normalization addressing the problem of vanishing feature activation. 
Our approach as well relies on the photometric loss and best practices in self-supervised optical flow prediction, but further exploits the availability of stereo image sequences to improve these concepts in the context of scene flow prediction.

\subsection{Self-Supervised Scene Flow}
The current research regarding scene flow prediction exhibits a large lack in self-supervised methods. The work of \cite{Hur2020} focuses on monocular scene flow prediction based on a photometric and 3D point reconstruction loss. However, the monocular input leads to a weakly posed problem, compared to the stereo setting. Furthermore, \cite{david2019flow} and \cite{zuanazzi2020adversarial} discuss self-supervision for point clouds. However, the sensing of point clouds is expensive and often results in an insufficient density. Due to these reasons, we solve -- for the first time -- the problem of self-supervised scene flow prediction in a stereo camera setting. 

\section{Self-Supervised Scene Flow Prediction} \label{sec:method}

\begin{figure}
     \centering
     \begin{subfigure}[c]{0.48\linewidth}
        \centering
         \begin{subfigure}[b]{\linewidth}
             \centering
             \includegraphics[width=\textwidth]{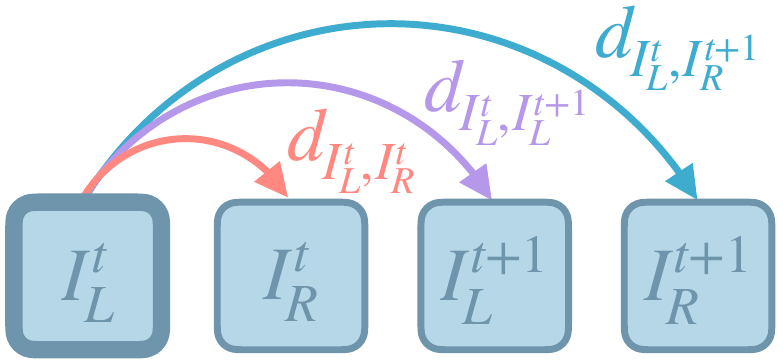}
             \caption{Flow prediction for reference image $\mathbf{I}_L^t$.}
             \label{fig:displace:forward}
         \end{subfigure}\\%
         \begin{subfigure}[b]{\linewidth}
             \centering
             \includegraphics[width=\textwidth]{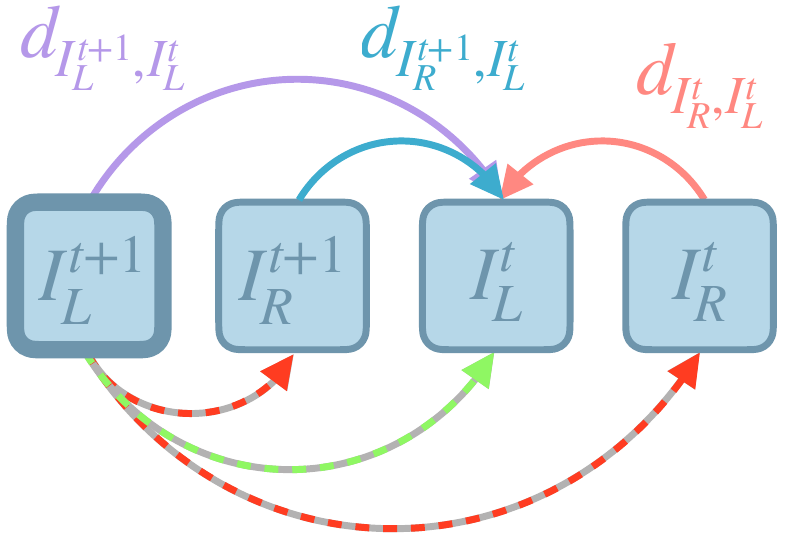}
             \caption{Flow prediction for reference image $\mathbf{I}_L^{t+1}$ and required inverse displacements for (\subref{fig:displace:forward}). The predictions in red can not be used as inverse displacements directly.}
             \label{fig:displace:backward_noflip}
         \end{subfigure}
     \end{subfigure}
     \hfill
     \begin{subfigure}[c]{0.48\linewidth}
         \centering
         \includegraphics[width=\linewidth]{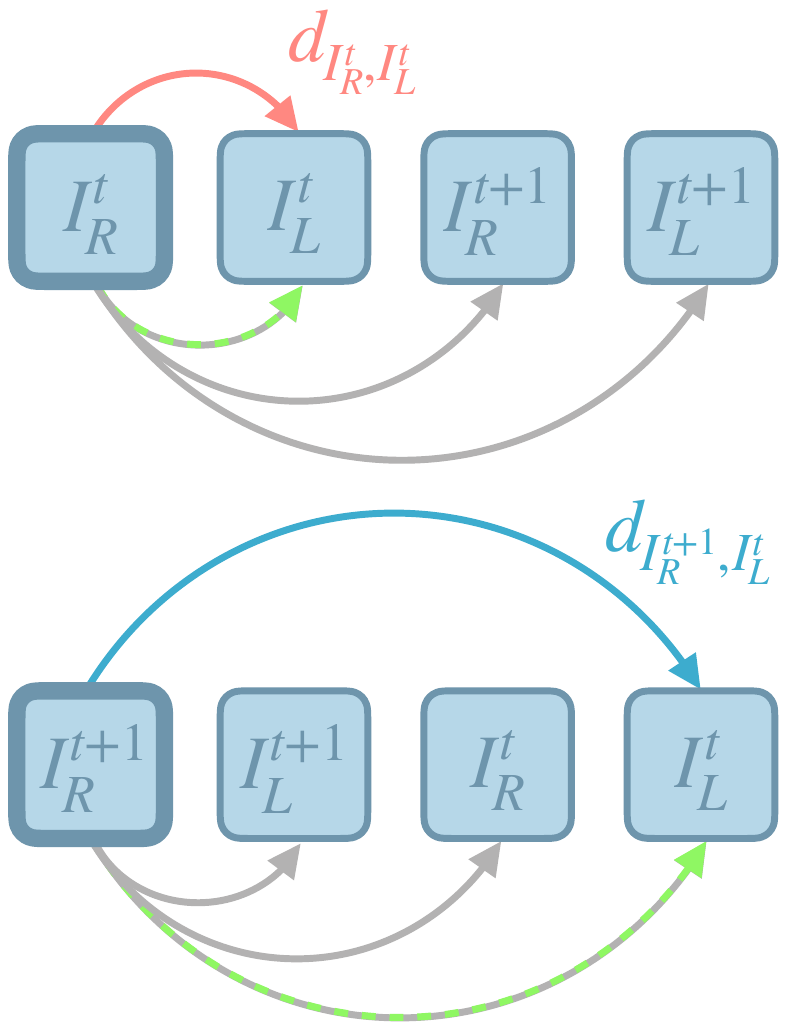}
         \caption{Flow predictions for reference images with spatial \textit{flipping}, which add the missing inverse displacements in (\subref{fig:displace:backward_noflip}).}
         \label{fig:displace:backward_flip}
     \end{subfigure}
        \caption{Visualization of the backward and forward displacement calculation with and without \textit{flipping}. Corresponding forward and backward displacement are colored in the same way. $d_{\mathbf{I}_1,\mathbf{I}_2}$ describes the displacement from image $\mathbf{I}_1$ towards $\mathbf{I}_2$.}
        \label{fig:displace}
\end{figure}

We assume a standard stereo setting for scene flow prediction, in which four images $\mathbf{I}_L^t, \mathbf{I}_R^t, \mathbf{I}_L^{t+1}, \mathbf{I}_R^{t+1}$ (left and right frames at two time steps) are consumed to produce a four dimensional scene flow prediction $\mathbf{s} = (u,v,disp_t, disp_{t+1})^T$ consisting of the optical flow field and two disparity maps \cite{Menze2015CVPR} (\cf \cref{fig:displace}).
By this, our self-supervised framework enables the training of a broad range of architectures for scene flow prediction on unlabelled data.
One deep neural network in compliance with this definition is PWOC-3D \cite{saxena2019pwoc3d}, which we choose as our base architecture, because of its simplicity and fast run time.
Additionally, our method leverages the stereo setup to compute bidirectional displacements for the estimation of occlusions. This way, the self-supervised loss can be easily applied symmetrically for multiple reference images.

\subsection{Self-Supervised Loss} \label{sec:method:loss}


Our self-supervised loss is based on the photometric reconstruction loss, \ie the difference between the reference and warped image based on the scene flow prediction. According to \cite{meister2018unflow}, this loss can be formulated, for two images $\mathbf{I}_1, \mathbf{I}_2$ which are related by the two dimensional displacement $d_{\mathbf{I}_1,\mathbf{I}_2}$, over all pixel positions $\mathbf{x}$ in the following way:
\begin{equation}
\begin{split}
     L(\mathbf{I}_1, \mathbf{I}_2) = &\\
      \sum_{\mathbf{x}} & \bigg( \mathbf{O}(\mathbf{x}) \cdot \rho\Big( \Phi\big( \mathbf{I}_1(\mathbf{x}), \mathbf{I}_2(\mathbf{x}+d_{\mathbf{I}_1,\mathbf{I}_2}(\mathbf{x})) \big)\Big)\\
     & + \lambda \big(1-\mathbf{O}(\mathbf{x})\big)\bigg),
\end{split}
\end{equation}
with $\rho(x)$ as the Charbonnier function \cite{Bruhn2005} and \sloppy $\Phi(\mathbf{I}_1(\mathbf{x}), \mathbf{I}_2(\mathbf{x}'))$ as the photometric difference at the corresponding pixel $\mathbf{x}$ and $\mathbf{x}'$. In order to introduce invariance for monotonic illumination changes, we apply the ternary census transform \cite{Zabih1994} on the images and compute the Hamming Distance.
Furthermore, we utilize a binary occlusion mask $\mathbf{O}$, in order to mask occluded pixels. We determine this occlusion mask by applying a consistency check \cite{sundaram2010} (see \cref{sec:method:occ}). To avoid the trivial solution of occluding everything, we add a constant penalty $\lambda=12.4$ for all occluded pixels \cite{meister2018unflow}. In addition, we use a second-order smoothness constraint \cite{Trobin2008prior}. To improve the self-supervision we also apply cost volume normalization \cite{jonschkowski2020matters} on the corresponding layer in our network as a means to support improved feature activation.
The reconstruction loss over two images is afterwards computed for all image pairs between the reference image $\mathbf{I}_L^t$ and the rest of the input images:
\begin{equation}
    L_{pairs}(\mathbf{I}_L^t)= L(\mathbf{I}_L^t, \mathbf{I}_R^t) + L(\mathbf{I}_L^t, \mathbf{I}_L^{t+1}) + L(\mathbf{I}_L^t, \mathbf{I}_R^{t+1}).
\end{equation}
 
\subsection{Bidirectional Displacements} \label{sec:method:occ}

The forward displacements from the reference image $\mathbf{I}^t_L$ towards all the other images are the network's prediction for the default input.
A bidirectional consistency check for the estimation of occlusions requires the computation of the inverse displacements, based on an altered input.
The corresponding standard approach in optical flow is using an alteration of the time-wise order of the input images.
In our case, this results in the following input for the network $\mathbf{I}_L^{t+1}, \textbf{ I}_R^{t+1}, \textbf { I}_L^{t}, \text{ and }\mathbf{I}_R^{t}$, in which $\mathbf{I}_L^{t+1}$ is the reference image.
To derive corresponding backward displacements for the forward scene flow prediction, additional warping operations become necessary (\cf \cref{fig:displace}).
These not only require a greater computational overhead but also introduce holes in the predictions as well as an additional interpolation error. 
In order to avoid this obstacle, we propose spatial \textit{flipping} of the input, \ie not only altering the temporal but also the spatial order (left-right).
This allows us to predict scene flows for all images and therefore directly retrieve the corresponding inverse displacements from a suitable flow (see \cref{fig:displace:backward_flip}).
Based on the predicted scene flows our approach is moreover easily extendable to consider losses for additional reference images, compared to supervised approaches, which only focus on the reference image $\mathbf{I}_L^{t}$. 
Favoring $\mathbf{I}_L^{t}$ as reference view has no conceptual reasoning, since optical challenges like occlusion, specular reflections, noise \etc can occur in all frames \cite{prsm}. Furthermore, multiple reference images provide more information and thus enforce further restrictions on the correctness of the scene flow prediction.
If multiple reference images $I_{ref}$ are considered, the overall loss consists of the sum of pairwise losses for all of them:
\begin{equation}
    L_{total} = \sum_{\mathbf{I} \in I_{ref}} L_{pairs}(\mathbf{I}).
\end{equation}

\section{Experiments and Results} \label{sec:experiments}

\subsection{Details} \label{sec:experiments:details}
Since our work focuses primarily on real world scenarios, the KITTI raw data set \cite{Geiger2013IJRR} is a natural choice for a training data set. Furthermore, we utilize the annotated KITTI 2015 data set \cite{Menze2015CVPR} for validation and fine-tuning using the same splitting as in \cite{saxena2019pwoc3d}. In order to ensure unbiased and comparable results, we remove sequences from the training data, that also appear in the validation data, resulting in 35666 training samples. In the fashion of \cite{saxena2019pwoc3d}, we further use the Adam \cite{kingma2015adam} optimizer and apply a learning rate of $0.00017$ with a batch size of $b=3$. For the loss we utilize the recommended parameter settings in \cite{meister2018unflow}. For all our experiments, we evaluate the average end-point error (EPE [px]) and the KITTI outlier error (KOE [\%]) \cite{Menze2015CVPR} of the predicted scene flow field.

\subsection{Ablation Study} \label{sec:experiments:ablation}

\begin{table}[t]
\centering
\caption{Ablation study for different components of our photometric loss on our validation set. We assess the influence of cost volume normalization, the choice of reference images, as well as the application of \textit{flipping}.}
\resizebox{\columnwidth}{!}{
\begin{tabular}{c c c c c c c}
\hline
    Costvol. Norm. & Ref. Image(s) & Flipping & EPE & KOE\\
    \hline
 \cmark  &    $I^t_L$  & \xmark  & 14.26 & 54.14\\
  \xmark & $I^t_L, I^t_R$  & \xmark    & 19.78 & 47.06\\ 
   \cmark    &    $I^t_L, I^t_R$  & \xmark    & \textbf{10.79} & 36.19   \\
   \cmark    &    $I^t_L, I^t_R$  & \cmark   & 12.22 & \textbf{30.35}      \\
  \cmark    &    $I^t_L, I^t_R, I^{t+1}_L, I^{t+1}_R$  & \cmark   & 11.93 & 31.32\\
  \hline
\end{tabular}
}
\label{tab:photo}
\end{table}

As evident from \cref{tab:photo}, additional reference images enhance the performance of the model due to the improved representation of the relationship between the predicted scene flow and the corresponding image pairs which enforces further constraints on the correctness of the prediction. However, if \textit{flipping} is applied its effect decreases, leading to similar results between 2 and 4 reference images due to the great amount of information already included in the loss. Furthermore, the results show, that the utilization of \textit{flipping} drastically improves the performance. This is caused by the supplementary information it offers during the training process which enhances the networks understanding of the relationship between bidirectional flows. We can also observe a clearly positive effect from applying cost volume normalization.

\subsection{Comparison to Supervised Pre-Training} \label{sec:experiments:self_vs_super}






\begin{figure*}
     \centering
     \begin{subfigure}[c]{0.24\textwidth}
         \centering
         \includegraphics[width=\textwidth]{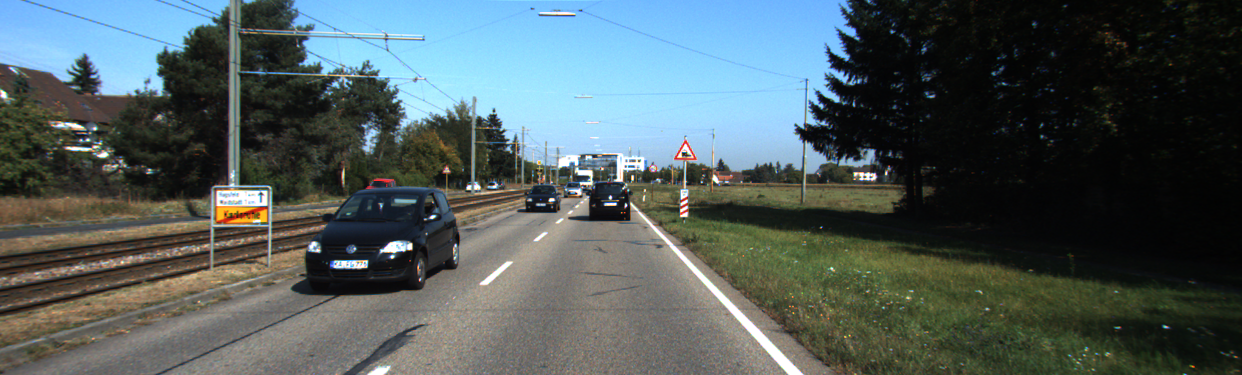}
         \caption{Reference Image $I^t_L$.}
         \label{fig:compare:img}
     \end{subfigure}
     \begin{subfigure}[c]{0.24\textwidth}
         \centering
         \includegraphics[width=\textwidth]{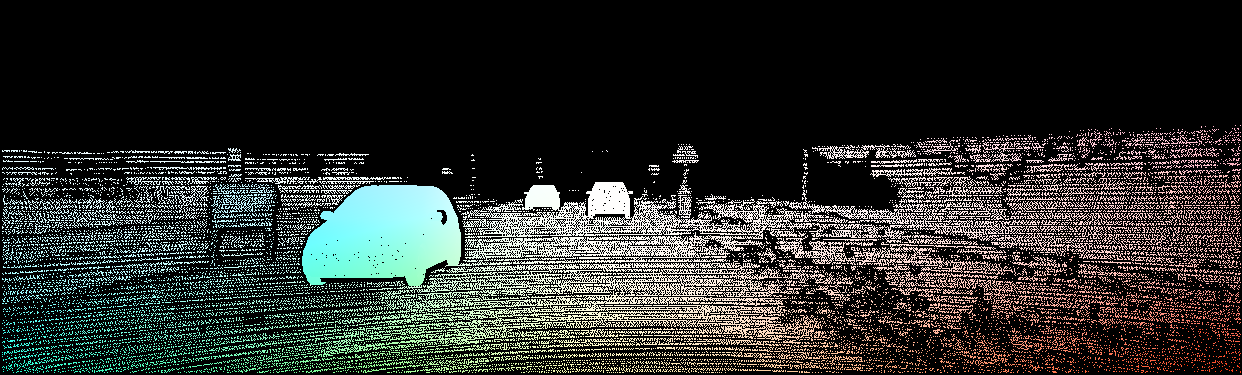}
         \includegraphics[width=\textwidth]{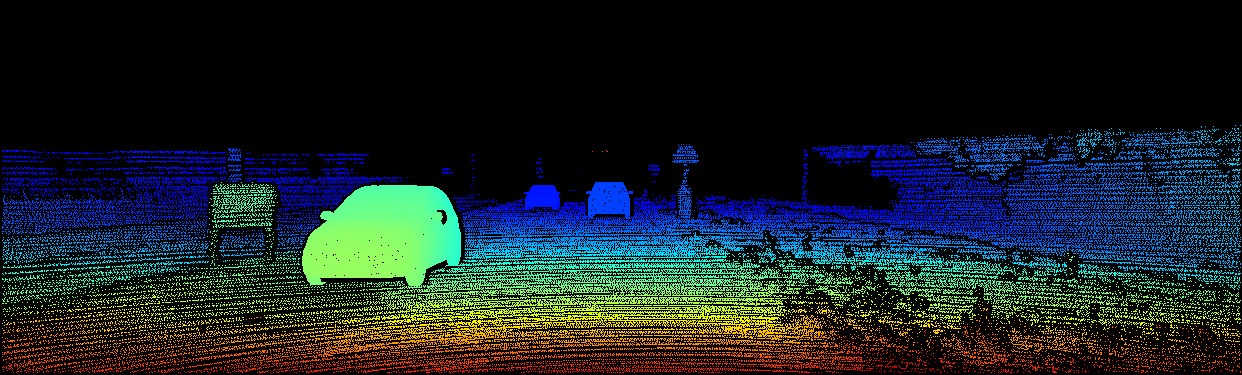}
         \caption{Ground Truth (visually enhanced).}
         \label{fig:compare:gt}
     \end{subfigure}
     \begin{subfigure}[c]{0.24\textwidth}
         \centering
         \includegraphics[width=\textwidth]{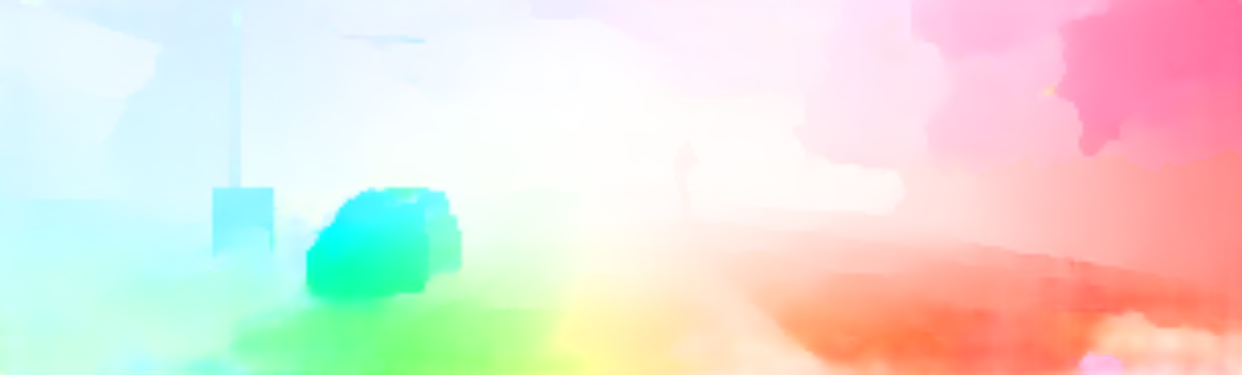}
         \includegraphics[width=\textwidth]{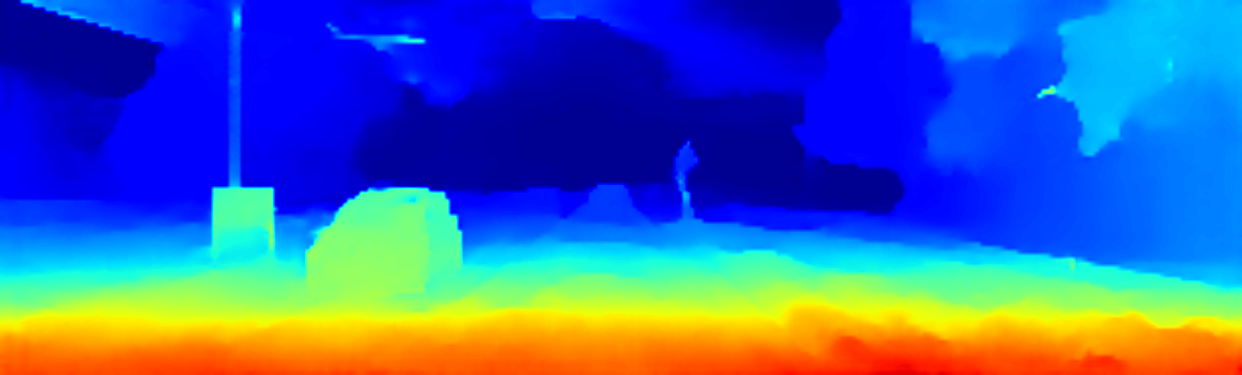}
         \caption{Supervised pre-training on synthetic data.}
         \label{fig:compare:super}
     \end{subfigure}
     \begin{subfigure}[c]{0.24\textwidth}
         \centering
         \includegraphics[width=\textwidth]{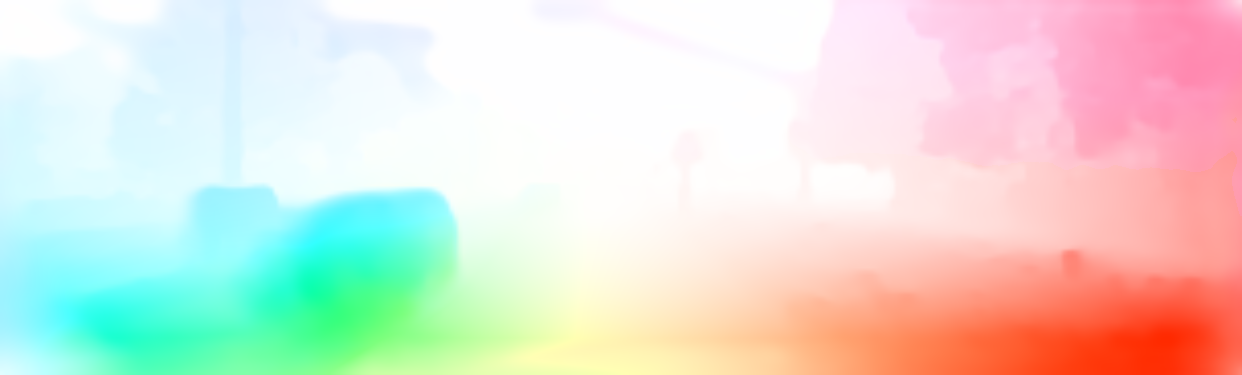}
         \includegraphics[width=\textwidth]{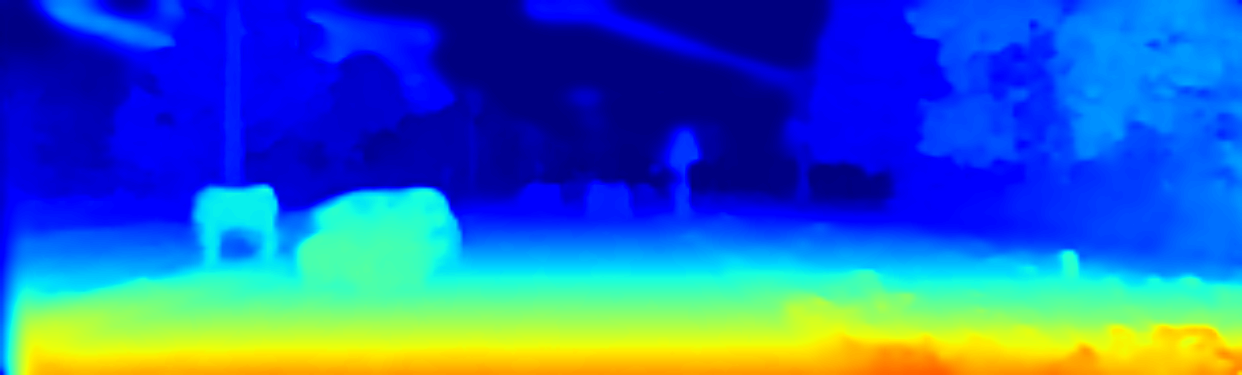}
         \caption{Pretrained with our Self-SuperFlow.}
         \label{fig:compare:photo}
     \end{subfigure}
     \caption{Visualization of optical flow and first disparity for the supervised pre-training and our photometric loss on a KITTI validation sample before fine-tuning.}
    \label{fig:compare}
\end{figure*}

\begin{table}[t]
\centering
\caption{Comparison between the supervised training method (as in PWOC-3D \cite{saxena2019pwoc3d}) and our self-supervised method on our validation split, before and after supervised fine-tuning.}
\resizebox{\columnwidth}{!}{
\begin{tabular}{c c c c c c}
\hline
    Data & Supervised & Stage & Epochs & EPE-all & KOE-all  \\ 
    \hline
    synthetic & \cmark & pre & 760  & 14.52 & 48.91    \\
    real & \xmark & pre  & 24 & \textbf{12.22} & \textbf{30.35}\\
    \hline
    real & \cmark & fine  & 125 & \textbf{3.52} & 13.76  \\
    real & \cmark & fine  & 125 & 3.92 &  \textbf{13.70}\\
    \hline
\end{tabular}
}
\label{tab:comp}
\end{table}

\Cref{tab:comp} shows that the photometric approach is able to drastically outperform the supervised pre-training on synthetic data with almost 20 percentage points less outliers and about 2px lower EPE. This indicates the strong influence of the domain gap between synthetic and real images. The KITTI raw data set \cite{Geiger2013IJRR} used for our self-supervised method resembles the target domain much closer and is further able to provide higher versatility and variation compared to the synthetic data used in the supervised pre-training.
A more detailed evaluation reveals that our self-supervised approach is inferior in terms of EPE in the foreground areas. The reason for this are occlusions, which lead to a lack of information that can not be overcome by the self-supervised method.
Furthermore, foreground areas generally pose a greater challenge since they typically involve higher magnitudes of displacements. 
After fine-tuning, the self-supervised training is still capable of achieving similar results compared to the supervised one.
\Cref{fig:compare} further shows, that the supervised method produces sharper edges, which may be hindered by the smoothness constraint in the self-supervised method. However it can be seen, that the photometric loss is able to capture more detail especially in background areas.
In addition, our self-supervised method requires only 24 epochs ($\sim 0.285 M$ iterations with batches of $3$) of pre-training, while the supervised approach requires 760 training epochs ($\sim 7.84 M$ iterations with batches of $2$). This validates the improved convergence capabilities when real world data is used for the training.

\subsection{Comparison to SotA on the KITTI Benchmark}
\label{sec:experiments:kitti}

\begin{table}[t]
\centering
\caption{Comparison of self-supervised and supervised methods on the KITTI test set (online benchmark).}
\resizebox{\columnwidth}{!}{
\begin{tabular}{c l c c c c}
\hline
 & Method  & D1-all  & D2-all & Fl-all & SF-all  \\ \hline
  \multirow{3}{*}{\rotatebox[origin=c]{90}{stereo}} & Self-SuperFlow-ft (Ours)  & \textbf{4.66} &  8.65 & \textbf{12.12} & \textbf{14.73}\\
& PWOC-3D-ft \cite{saxena2019pwoc3d}  & 5.13 & \textbf{8.46} & 12.96  & 15.69    \\
  & Self-SuperFlow (Ours)  & 8.11  & 21.57 & 23.67 & 28.71\\
  \hline
 \multirow{4}{*}{\rotatebox[origin=c]{90}{mono}} & Multi-Mono-SF-ft \cite{hur2021self}   &	22.71&		26.51&		13.37&	33.09\\
   & Self-Mono-SF-ft  \cite{Hur2020} &	22.16&	25.24&	15.91&	33.88\\
   & Multi-Mono-SF \cite{hur2021self}  & 30.78  & 34.41 & 19.54 & 44.04 \\
   & Self-Mono-SF \cite{Hur2020} & 34.02 & 36.34 & 23.54 & 49.54	\\
  \hline
\end{tabular}
}
\label{tab:comp_test}
\end{table}

\Cref{tab:comp_test} shows the results of our method compared to PWOC-3D and other self-supervised approaches on the KITTI scene flow benchmark\footnote{\url{http://www.cvlibs.net/datasets/kitti/eval_scene_flow.php}}. It is evident, that after fine-tuning our approach achieves a superior overall scene flow outlier rate compared to the other ones. It even outperforms the supervised training of PWOC-3D by 1 \%, showing improved generalization capabilities after fine-tuning when real world data is used for the pre-training. Especially the result in the disparity prediction surpasses all the other methods.
To no surprise, our method cuts the outlier rates by half compared to the monocular methods Self-Mono-SF \cite{Hur2020} and Multi-Mono-SF \cite{hur2021self}.

\section{Conclusion} \label{sec:conclusion}
In this work we approach the lack of real world ground truth data and the inherent representation gap due to the usage of synthetic training images by introducing a novel self-supervised training method for scene flow prediction in stereo sequences, which can be applied to a vast range of recent networks. Our approach is based on a bidirectional reconstruction loss as well as a forward-backward consistency check for occlusion awareness. We further propose an innovative strategy to derive the inverse displacement in a scene flow setting using \textit{flipping} \ie spatial variations of the input images. 

Without fine-tuning, our self-supervised approach outperforms the equivalent supervised training by over 20 percentage points less outliers on the validation set, proving the significant influence of the representation gap on the training performance. We were further able to show a faster convergence during training and an improved generalization capability of our method compared to the supervised one.

\bibliographystyle{IEEEbib}
\bibliography{ref}

\end{document}